\pdfoutput=1

\documentclass[11pt]{article}

\usepackage[final]{acl}

\usepackage{times}
\usepackage{latexsym}
\usepackage{enumitem}
\usepackage[T1]{fontenc}

\usepackage[utf8]{inputenc}

\usepackage{microtype}

\usepackage{inconsolata}

\usepackage{graphicx}

\usepackage{soul}
\usepackage{url}
\usepackage{amsmath}
\usepackage{amsthm}
\usepackage{booktabs}
\usepackage{algorithm}
\usepackage{algpseudocode}
\usepackage[switch]{lineno}
\usepackage{multirow} 
\usepackage{subcaption}
\usepackage{pifont}
\usepackage{xcolor}
\usepackage{caption}
\title{From Generation to Detection: A Multimodal Multi-Task Dataset\\ for Benchmarking Health Misinformation}

\author{Zhihao Zhang\textsuperscript{1}\thanks{Equal contribution}, Yiran Zhang\textsuperscript{1}\footnotemark[1], Xiyue Zhou\textsuperscript{2}, Liting Huang\textsuperscript{3},\\ \textbf{Imran Razzak}\textsuperscript{4}, \textbf{Preslav Nakov}\textsuperscript{4}, \textbf{Usman Naseem}\textsuperscript{1} \\
 Macquarie University\textsuperscript{1},  University of Sydney\textsuperscript{2}, UTS\textsuperscript{3}, MBZUAI\textsuperscript{4} \\
\texttt{\{zhihao.zhang, yiran.zhang, usman.naseem\}@mq.edu.au} \\
\texttt{\{imran.razzak, preslav.nakov\}@mbzuai.ac.ae
}}

\begin{document}
\maketitle
\begin{abstract}
Infodemics and health misinformation have significant negative impact on individuals and society, exacerbating confusion and increasing hesitancy in adopting recommended health measures. Recent advancements in generative AI, capable of producing realistic, human-like text and images, have significantly accelerated the spread and expanded the reach of health misinformation, resulting in an alarming surge in its dissemination. To combat the infodemics, most existing work has focused on developing misinformation datasets from social media and fact-checking platforms, but has faced limitations in topical coverage, inclusion of AI-generation, and accessibility of raw content. To address these gaps, we present MM-Health, a large scale multimodal misinformation dataset in the health domain consisting of 34,746 news article encompassing both textual and visual information. MM-Health includes human-generated multimodal information (5,776 articles) and AI-generated multimodal information (28,880 articles) from various SOTA generative AI models. Additionally, We benchmarked our dataset against three tasks—reliability checks, originality checks, and fine-grained AI detection—demonstrating that existing SOTA models struggle to accurately distinguish the reliability and origin of information. Our dataset aims to support the development of misinformation detection across various health scenarios, facilitating the detection of human and machine-generated content at multimodal levels\footnote{Our code and data is available at: \url{https://github.com/grantzyr/MM-Health-Dataset}}.
\end{abstract}
\section{Introduction}
Health misinformation refers to information that is inaccurate, misleading, or false according to the best available health evidence at the time. Such misinformation tends to spread at unprecedented speed and scale on the World Wide Web and social media~\cite{Office_of_the_Surgeon_General_OSG2021-op}, and has been shown to have significant negative effects on society~\cite{Borges_do_Nascimento2022-nh, Abbott2021-bg, WANG2019112552, Muhammed_T2022-wu, walter-2020}. During crises, such as outbreaks of infectious diseases like COVID-19, the overproduction of health misinformation in both digital and physical environments is defined as an infodemic. As society transitions into a long COVID era~\cite{Davis2023-qm}, the infodemic continues to evolve, leading to a reduction in public trust in health professionals~\cite{walter-2020}. Furthermore, unfiltered exposure to health misinformation can delay or prevent effective disease treatment and even threaten the lives of individuals~\cite{WANG2019112552}. Additionally, the recent advent of generative image models, such as Stable Diffusion~\cite{stable-diffusion}, MidjourneyV5~\cite{midjourney}, DALL-E2~\cite{dall-e}, as well as human-level text generators like ChatGPT~\cite{chat-gpt} and LLaMa~\cite{llama}, has amplified both the quantity and spread of misinformation~\cite{10.1145/3544548.3581318}. The ease with which generative AI models can replicate or manipulate multimodal content, including text and images~\cite{huang2024ruailargemultimodaldataset}, further contributes to the emergence of health misinformation~\cite{Park2024-ma,ahmad2025vaxguard,shah2024navigating}. This presents a significant new challenge in combating the infodemic.

\begin{table*}[ht]
\centering
\resizebox{0.8\textwidth}{!}{%
\begin{tabular}{lccccccccc}
\toprule
\multirow{2}{*}{\textbf{Name}} & \multirow{2}{*}{\textbf{Year}} & \multicolumn{2}{c}{\textbf{Human Context}} & \multicolumn{2}{c}{\textbf{Machine Context}} & \multirow{2}{*}{\textbf{\begin{tabular}[c]{@{}c@{}}Multiple \\ AI Models\end{tabular}}} & \multirow{2}{*}{\textbf{Reliability}} & \multirow{2}{*}{\textbf{Originality}} & \multirow{2}{*}{\textbf{\begin{tabular}[c]{@{}c@{}}General \\ Avaliable\end{tabular}}} \\ \cmidrule{3-6}
                               &                                & \textbf{Text}          & \textbf{Image}         & \textbf{Text}          & \textbf{Image}         &                                                                                         &                                       &                                       &                                                                                        \\ \midrule
MedHelp                        & 2013                           & \ding{52} & \ding{56} & \ding{56} & \ding{56} & \ding{56}                                                                  & \ding{52}                & \ding{56}                & \ding{56}                                                                 \\
COAID                          & 2020                           & \ding{52} & \ding{56} & \ding{56} & \ding{56} & \ding{56}                                                                  & \ding{52}                & \ding{56}                & \ding{56}                                                                 \\
ANTi-Vax                       & 2021                           & \ding{52} & \ding{56} & \ding{56} & \ding{56} & \ding{56}                                                                  & \ding{52}                & \ding{56}                & \ding{56}                                                                 \\
MM-COVID                       & 2020                           & \ding{52} & \ding{52} & \ding{56} & \ding{56} & \ding{56}                                                                  & \ding{52}                & \ding{56}                & Partly                                                                                 \\
ReCOVery                       & 2020                           & \ding{52} & \ding{52} & \ding{56} & \ding{56} & \ding{56}                                                                  & \ding{52}                & \ding{56}                & Partly                                                                                 \\
Monant                         & 2022                           & \ding{52} & \ding{52} & \ding{56} & \ding{56} & \ding{56}                                                                  & \ding{56}                & \ding{56}                & Partly                                                                                 \\
MMCOVAR                        & 2021                           & \ding{52} & \ding{52} & \ding{56} & \ding{56} & \ding{56}                                                                  & \ding{52}                & \ding{56}                & Partly                                                                                 \\
Med-MMHL                       & 2023                           & \ding{52} & \ding{52} & \ding{52} & \ding{56} & \ding{56}                                                                  & \ding{52}                & \ding{52}                & Partly                                                                                 \\ \midrule
Ours                           & 2025                           & \ding{52} & \ding{52} & \ding{52} & \ding{52} & \ding{52}                                                                  & \ding{52}                & \ding{52}                & \ding{52}                                                                 \\ \bottomrule
\end{tabular}
}
\caption{Comparison between our MM-Health dataset and other health related misinformation datasets.}
\label{tab:data}
\end{table*}

Recent studies have focused on developing and analyzing datasets and detection methods to combat the infodemic. Common methods for collecting multimodal health misinformation include scraping social media platforms~\cite{MedHelp, ANTi-Vax, CoAID, ReCOVery, Monant, Med-MMHL, MM-COVID, MMCoVaR}, such as Twitter (now known as X), and parsing fact-checking websites~\cite{MM-COVID, MMCoVaR, Monant, Med-MMHL}, such as Snopes and Poynter. 
However, these datasets exhibit the following notable limitations:
\begin{itemize}[noitemsep,leftmargin=*]
    \item \textbf{Exclusion of AI-generated misinformation:} The majority of existing datasets focus solely on human-generated misinformation, ignoring the growing trend of AI-generated misinformation~\cite{Synthetic, 10.1145/3687028, 10.1145/3581783.3612704}. With AI models now capable of generating and manipulating both text and image content~\cite{LIZLOPEZ2024102103}, it is crucial to include AI-generated misinformation to ensure data diversity.
    \item \textbf{Lack of accessible raw content:} Many datasets only provide URLs or Twitter IDs as delivery artifacts instead of the raw content of the misinformation. Due to ongoing censorship by news and social media platforms, much of this data is no longer accessible, severely limiting the usability of these datasets.
\end{itemize}

The limitation of the previous health misinformation datasets are summarised in Table~\ref{tab:data}. Most of the existing methods for automatically detecting health misinformation include content-based CNNs~\cite{CoAID, ReCOVery, DBLP:conf/icwsm/DaiSW20}, attention-based hierarchical encoders~\cite{Kinkead2020}, and graph-based attention methods~\cite{DETERRENT}. These methods have demonstrated competitive performance after being trained on health misinformation benchmarks. However, they require extensive training on benchmark datasets to achieve expected performance, which limits the models' ability to generalise to new health misinformation. Since health misinformation is constantly evolving~\cite{Okoro2024ARO}, it is important to evaluate generalised models under zero-shot settings. Additionally, these models only accept structured input and provide labelled output without human-readable explanations. This constraint limits the usability of existing detection methods for the general public outside of the research community. To address the aforementioned limitations, we develop MM-Health, a comprehensive multimodal dataset designed for detecting both human and AI generated health misinformation. Additionally, we conduct extensive experiments using state-of-the-art (SOTA) Vision-Language Models (VLLMs) to evaluate their performance in assessing both the reliability and originality of the health information in MM-Health, as well as performing a fine-grained AI detection analysis. Our \textbf{key contributions} are as follows:
\begin{itemize}[noitemsep,leftmargin=*]
    \item We create and release a new large-scale multimodal (text and images) dataset called MM-Health, which contains 34,746 news articles. These articles were collected from existing multimodal health datasets and generated using state-of-the-art (SOTA) AI generative models.
    \item We conduct a preliminary analysis of the data for both human and AI-generated content, establishing our dataset as a benchmark for baseline evaluation against several SOTA Vision Large Language Models (VLLMs) across three tasks: reliability checks, originality checks, and fine-grained AI detection.
    \item Our experimental results and analysis show that current VLLMs, including GPT-4o~\footnote{\url{https://openai.com/index/hello-gpt-4o/}}, struggle to accurately verify content reliability and originality. This highlights the importance of developing generalised models for multimodal health misinformation detection. 
\end{itemize}

\section{Related Work}
Web and social media provide valuable sources of information and play important roles in multiple tasks like depression and suicide identification~\cite{DBLP:journals/tcss/NaseemKKD24}, health mention classification~\cite{DBLP:conf/www/NaseemKKD22} and creditable knowledge management~\cite{NISAR2019264}. However, systematic reviews shows that health-related misinformation on web and social media posts critical threat to the community, which could lead to delay or prevent treatment and even threaten the lives of individuals~\cite{WANG2019112552}.

\subsection{Existing Datasets}
As detailed in Table~\ref{tab:data}, several benchmark datasets have been proposed for health related information detection, which contain various features from different data sources. 

\noindent \textbf{MedHelp}~\cite{MedHelp}, \textbf{COAID}~\cite{CoAID}, and \textbf{ANTi-Vax}~\cite{ANTi-Vax} are text datasets designed to address online health misinformation. Specifically, MedHelp is collected from an online general health discussion forum. It contains 1,338 non-misinformation comments and 887 misinformation comments annotated by human experts between 2001 and 2013. COAID includes both news articles and Twitter content focused on COVID-19 between December 2019 and September 2020. It labels 204 fake news articles, 3,565 true news articles, 28 fake claims, and 454 true claims from a total of 1,896 news articles, 516 Twitter posts, and 183,569 Twitter engagements. ANTi-Vax is built solely from Twitter content focused on the COVID-19 vaccine between December 2020 and January 2021. It includes a total of 15,073 tweets, 5,751 of which are misinformation and 9,322 are general vaccine-related tweets.

\begin{figure*}[ht]
  \centering
  \includegraphics[width=0.8\textwidth]{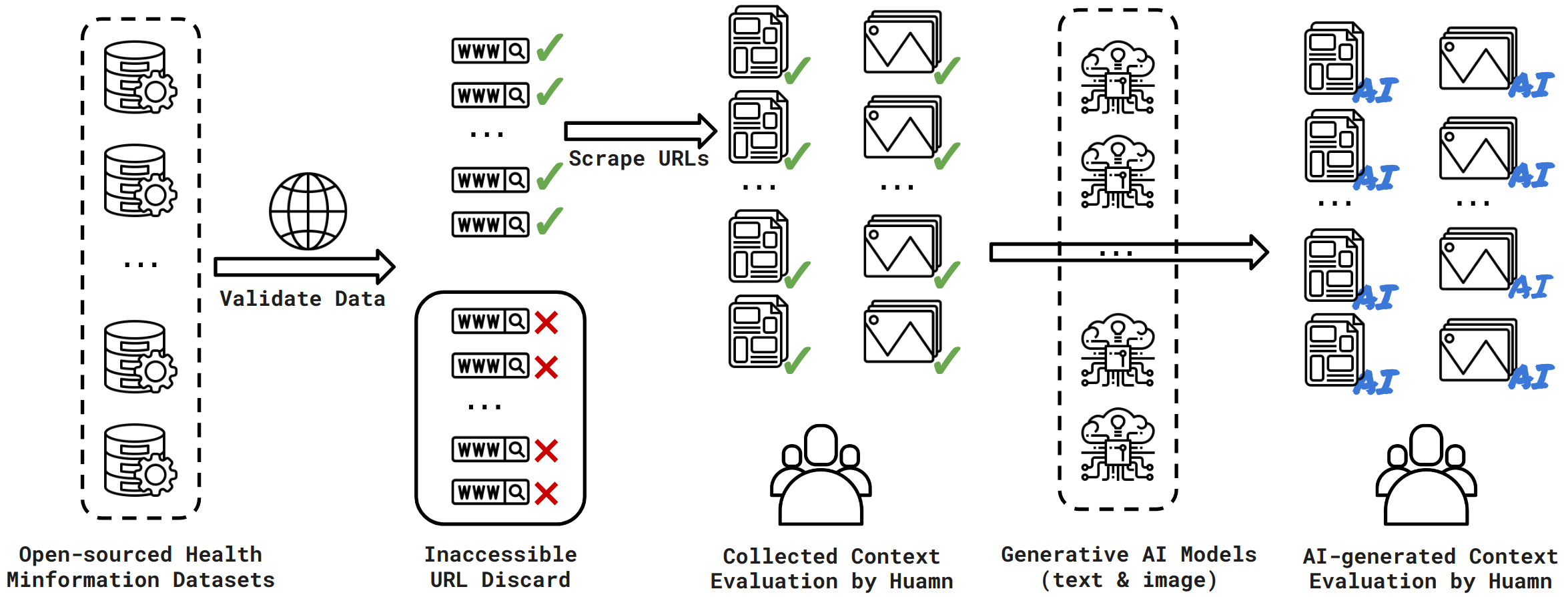}
  \caption{Data collection process for MM-Health includes: 1) utilising multiple existing open-source health misinformation datasets as data sources, 2) validating the available data samples and collecting human-generated multimodal data from the provided URLs, and 3) implementing generative AI models to collect AI-generated replicated multimodal data. To ensure data quality, both human and AI generated content are evaluated by five human evaluators proficient in English.}
  \label{fig:data-collection}
\end{figure*}

\noindent \textbf{MM-COVID}~\cite{MM-COVID}, \textbf{ReCOVery}~\cite{ReCOVery}, \textbf{Monant}~\cite{Monant}, \textbf{MMCoVaR}~\cite{MMCoVaR}, and \textbf{Med-MMHL}~\cite{Med-MMHL} are multimodal datasets for health-related misinformation detection. MM-COVID contains 3,981 fake articles and 7,584 real articles related to COVID-19 which collected from Feb 2020 to Jul 2020. ReCOVery includes both news articles and tweets mentioning COVID-19, labelled by NewsGuard~\footnote{\url{https://www.newsguardtech.com/}} and Media Bias/Fact Check~\footnote{\url{https://mediabiasfactcheck.com/}}. It comprises 118,339 reliable articles and 28,274 unreliable articles sourced between January 2020 and May 2020. MMCoVaR focuses on vaccine-related misinformation between February 2020 and March 2021, featuring 958 unreliable and 1,635 reliable news articles, plus 24,184 tweets categorized as reliable, unreliable, or inconclusive. Med-MMHL covers multiple diseases, integrating LLM-generated fake news, with 1,979 real and 1,604 fake articles, alongside 1,334 reliable and 639 unreliable tweets from January 2017 to May 2023. There are no existing dataset that incorporate both text and image AI for misinformation generation. Furthermore, the most datasets rely on website URLs or tweet IDs, which have become partially or fully inaccessible due to Twitter’s API restrictions and ongoing content censorship.

\subsection{Existing Methods}
\subsubsection{Methods for Textual Data} 
Existing methods typically utilize a latent space to create embedding representations for article content. Embeddings capture contextual information and encode it into a lower-dimensional representation for downstream tasks such as classification and prediction~\cite{papanikou2024healthmisinformationsocialnetworks}. Many approaches leverage this feature using the BERT to achieve exceptional performance in health misinformation classification, either through fine-tuning~\cite{FakeBERT} or a combination of pre-training and fine-tuning~\cite{BioBERT}. Other strategies, such as BERTweet~\cite{bertweet} and COVID-Twitter-BERT~\cite{COVID-Twitter-BERT}, combine multiple pre-trained models across various contexts. These hybrid models achieve higher performance due to the distinct contextual understanding provided by different models. However, textual models capture only single-modality features, neglecting the visual information often associated with articles. We address this limitation by using SOTA vision-language models to capture both textual and visual context.
\subsubsection{Methods for Multimodal Data}
Previous studies have found that image-containing posts or articles tend to have higher user interaction, spread more quickly, and trigger stronger emotional responses~\cite{papanikou2024healthmisinformationsocialnetworks}. Multimodal misinformation detection requires both image and text encoders to capture textual and visual features. One method~\cite{Uppada2023} leverages these encoders to identify fake posts containing visual data, while another approach detects inconsistencies between textual entities and image content for misinformation detection. Limited research has explored multimodal health misinformation detection. DGExplain~\cite{10.1145/3485447.3512257} is a recent multimodal encoder-decoder architecture for COVID-19 misinformation detection that investigates cross-modal associations between text and images to assess the reliability of news articles. However, these models often require extensive training on benchmark datasets and offer limited interpretability. Our work applies current advanced Vision-Language Models as a baseline for health misinformation detection to evaluate their performance.

\section{MM-Health Dataset}

The overall data collection process for MM-Health, including human and AI-generated content, is illustrated in Figure~\ref{fig:data-collection}. This process can be summarised as follows: 1) utilising multiple existing open-source health misinformation datasets as data sources, 2) validating the available data samples and collecting human-generated multimodal data from the provided URLs, and 3) implementing generative AI models to collect AI-generated replicated multimodal data. To ensure data quality, both human and AI generated content are evaluated by five human evaluators proficient in English.

\subsection{Data Collection}
\subsubsection{Health Misinformation Collection}
\label{sec:human_eval}
To ensure variability, we utilise four recent health-related misinformation datasets~\cite{Med-MMHL,MMCoVaR,MM-COVID,ReCOVery}, collected between February 2020 and May 2023, as our primary data sources. Since Twitter (now known as X.com) restricted data access through its application programming interface (API) in 2023, we exclude tweets from these datasets and focus on the available news articles. While Med-MMHL provides raw text and image data, the other datasets only provide URLs as their data source. However, not all URLs are accessible due to news websites consistently removing outdated or misleading news, either voluntarily or in response to complaints. Therefore, we validate the URLs in the database and scrape the text and image metadata from the available URLs and attach the reliable and unreliable labels extracted from the original datasets. We also use the Pillow\footnote{\url{https://python-pillow.org/}} library to filter out  some images, such as blurry, logo-based, or meaningless images, which are often downloaded during web scraping. Examples of removed images are in Appendix (Figure \ref{fig:remove-images}).

\subsubsection{AI Health Misinformation Generation}
After collecting human generated misinformation from open-source datasets, we deploy five state-of-the-art (SOTA) generative AI models for each modality to generate AI replicated data. To ensure data variability, we adopt text and image models with different architectures. For text generation, we use Llama-3.1-8B, Qwen2.5-7B, ChatGLM-4-9B, Gemma2-9B, and Mistral-v0.3-7B. For image generation, we use FLUX.1-dev~\footnote{\url{https://github.com/black-forest-labs/flux}}, Stable Diffusion 1.5, Stable Diffusion XL, Stable Diffusion VAE, and Stable Diffusion PAG. To enhance model understanding during text generation, we integrate both textual and visual context, ensuring diverse and comprehensive outputs. Specifically, we implement a two-step generation pipeline as follows: 1) providing raw text and image data to GPT-4o to create a topic summary for the article, and 2) inputting the topic summary into the five selected LLMs to generate the desired text. We find that specific and concise prompts yield the best results for LLMs. Image generation follows a similar two-step pipeline to that of text generation. First, the images are sent to GPT-4o to generate short image captions. Next, the original images and their captions are supplied to the image generation models to create new images. The original images serve as anchor points during the generation process, predefining the de-noising parameters as well as the image dimensions. All the text and image models are provided with their default parameter settings. Refer to Figure~\ref{fig:prompts} in appendix for text generation prompts and image generation prompts.


\subsubsection{Post-Processing with Human Evaluation}

Our two-stage generation process allows AI models to create most of the text and images based on existing data without triggering their internal censorship mechanisms~\cite{glukhov2023llmcensorshipmachinelearning}. However, some data samples are still not generated, as models may refuse to produce them or output empty strings and black images. Since different models are equipped with different censorship mechanisms, there are variations in behaviour between models. To address this, we developed an algorithm to post-process data alignment between the human-generated context and all five AI-generated contexts(see the Algorithm~\ref{alg:consensus_align} in appendix for the implementation ). The dataset consists of two sources: human-collected text and images from open-source datasets and AI-generated text-image pairs created by five different models. For consistency, our algorithm will ensured that each AI-generated sample was present across all five models before being included in the final dataset. After data matching, human evaluation was conducted by five independent annotators. These annotators were students ranging from junior to senior levels, including both male and female participants. All had backgrounds in AI literature and a foundational understanding of health-related topics, which helped ensure informed and thorough evaluation. Before the annotation task, all participants underwent thorough training to familiarize them with the evaluation criteria, common failure modes (e.g., logo artifacts, mismatched captions), and the overall goal of the task. This training ensured consistency and reliability in their assessments. The annotators were compensated fairly according to the university’s standard participant payment guidelines. Inter-annotator agreement was measured and found to be high, with a score above 80\%, indicating strong consistency across evaluators. While domain-specific expertise was not strictly required, the annotators’ familiarity with AI and health contexts contributed to a more nuanced understanding of the generated image–text pairs. This results in 34,746 pieces of health-related information, of which 5,791 were collected from open-source datasets and 28,955 were generated by AI models Examples of reliable and unreliable news generated by humans and AI are in Figure~\ref{fig:data-sample} in the appendix. The news articles are written in diverse styles by different LLMs while addressing the same topic. While Stable Diffusion 1.5 generates dramatically different images, other models slightly manipulate the original images while maintaining a realistic appearance.

\begin{table}[t]
\centering
\begin{minipage}{0.26\textwidth}
\centering
\resizebox{\textwidth}{!}{%
\begin{tabular}{@{}ccc@{}}
\toprule
\multicolumn{1}{c}{\textbf{Model}} & \multicolumn{1}{c}{\textbf{Avg Len}} & \multicolumn{1}{c}{\textbf{Avg CSS}} \\ \midrule
\textbf{Llama-3.1-8B}              & 479.57                     & 0.761                        \\
\textbf{Qwen2.5-7B}                & 568.4                      & 0.766                        \\
\textbf{ChatGLM-4-9B}              & 577.55                     & 0.761                        \\
\textbf{Gemma2-9B}                 & 340.45                     & 0.774                        \\
\textbf{Mistral-v0.3-7B}           & 337.87                     & 0.783                        \\ \bottomrule
\end{tabular}%
}
\end{minipage}%
\hfill
\begin{minipage}{0.21\textwidth}
\centering
\resizebox{\textwidth}{!}{%
\begin{tabular}{@{}ccc@{}}
\toprule
\multicolumn{1}{c}{\textbf{Model}} & \multicolumn{1}{c}{\textbf{FID}} & \multicolumn{1}{c}{\textbf{Avg CIS}} \\ \midrule
\textbf{FLUX.1-dev}                & 12.33                   & 0.869                        \\
\textbf{SD 1.5}      & 27.72                   & 0.737                        \\
\textbf{SD XL}       & 19.42                   & 0.866                        \\
\textbf{SD XL VAE}   & 15.76                   & 0.883                        \\
\textbf{SD XL PAG}   & 19.30                   & 0.866                        \\ \bottomrule
\end{tabular}%
}
\end{minipage}
\caption{Statistics of generated text and image data from different models. The left table shows text statistics: "Avg Len" is average word count, and "Avg CSS" is average Cosine Semantic Similarity. The right table shows image statistics: "FID" is Fréchet Inception Distance and "Avg CIS" is Average Cosine Image Similarity.}
\label{tab:side-by-side}
\end{table}

\subsection{Data Statistics and Analysis}
\noindent \textbf{Overall data statistics.} All instances in MM-Health are multimodal, containing both textual and visual information for reliability and originality studies, with the proportion of reliable to unreliable articles being approximately 4:1. We split our dataset into training and testing sets by randomly selecting 80\% and 20\% of the news articles from the collected data instances. Additionally, 10\% of the training set is further split into a validation set. As shown in Table~\ref{tab:overall-statistic}, this results in a total of 4,154 training samples, 463 validation samples, and 1,159 testing samples from human-generated news articles, while 20,770, 2,315, and 5,795 samples are used for training, validation, and testing, respectively, from AI-generated news articles. To ensure the generalisation of our dataset across all split sets, we further analyse the data statistics in each set, including the average text length and the average number of images per article from both human and AI sources. The average text length and number of images are consistent across different sets. It is worth noting that the average text length generated by AI is shorter than that of the original human text, with an average of around 450 words compared to approximately 850 words. Additionally, unreliable news articles tend to be shorter, averaging around 650 to 750 words, while containing more images per article around 6 to 8 images compared to around 1.5 images in reliable articles.The overall statistics of the dataset are presented in appendix Table~\ref{tab:overall-statistic}.

\begin{figure}[t]
  \centering
  \includegraphics[width=0.35\textwidth]{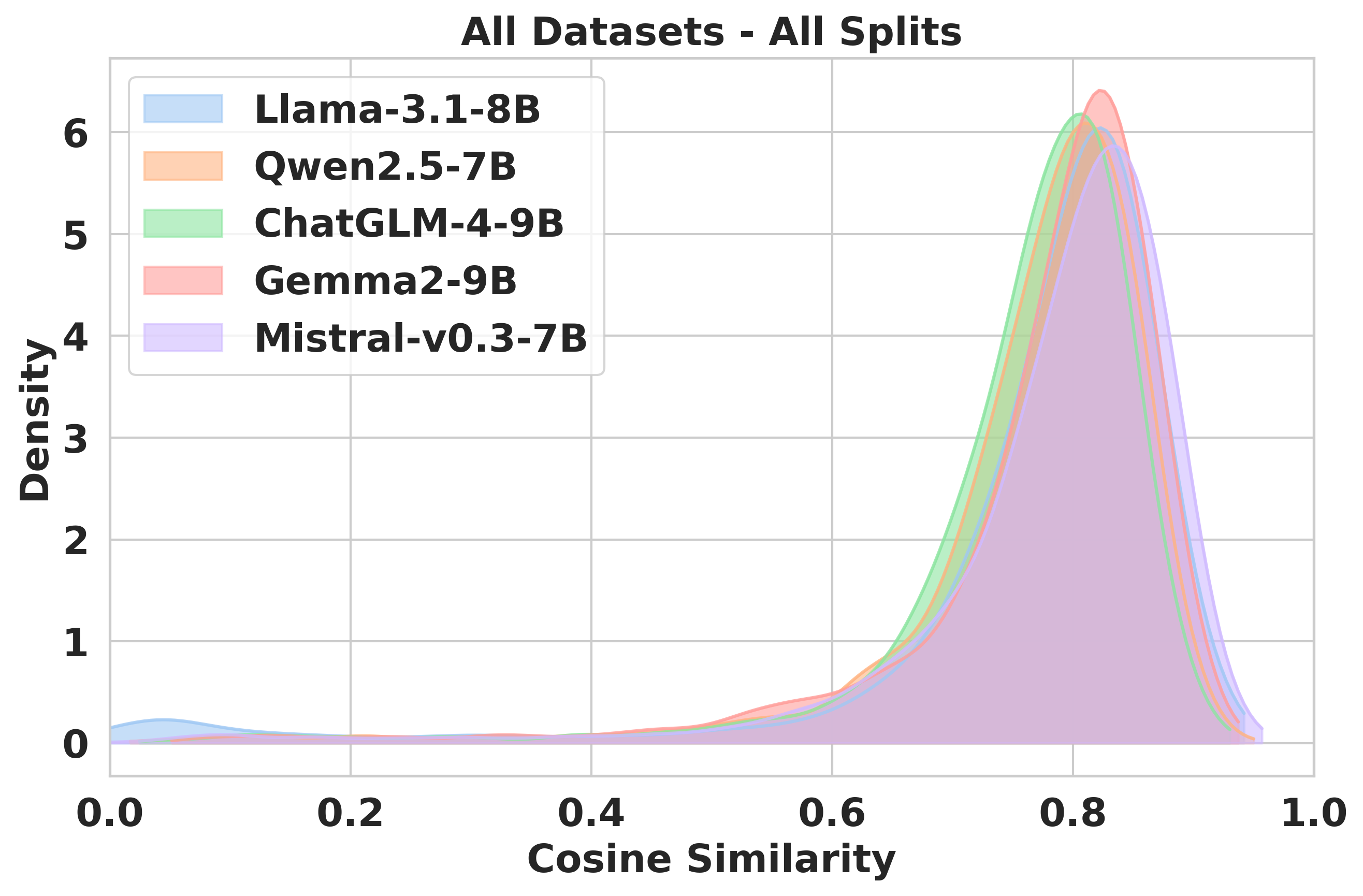}
  \caption{KDE distribution of the semantic similarity between the human articles and articles from five LLMs.}
  \label{fig:text-density}
\end{figure}

\noindent \textbf{Data analysis.} To identify the differing data characteristics across the five AI models, we compare their statistics in Table~\ref{tab:side-by-side}. For text models, we calculated the average article length from all five text models. There is a noticeable variation among the models, with ChatGLM-4 tending to generate longer articles, while Mistral-v0.3 is more likely to generate shorter articles. Additionally, we utilise the OpenAI \texttt{text-embedding-3-small} model to generate text embeddings for both human and AI articles, comparing their semantic differences using cosine similarity. For news articles that are too long to fit into the model's 8,191 context window, we split these long articles into smaller chunks and compute the mean vectors of all chunks to obtain the final text embedding representations for those articles. As indicated on the left side of Table~\ref{tab:side-by-side}, the average cosine semantic similarities across the text models range from 0.76 to 0.78. We plot the Gaussian Kernel Density Estimation (KDE)~\cite{DBLP:journals/isci/ZhuLHK22} distribution of the semantic similarity in Figure~\ref{fig:text-density}, showing that the cosine similarities are concentrated between 0.8 and 0.85 across all models, with slight variance. Notably, a small number of generated articles exhibit low semantic similarity when compared to human generated articles. 

\begin{figure}[t]
  \centering
  \includegraphics[width=0.35\textwidth]{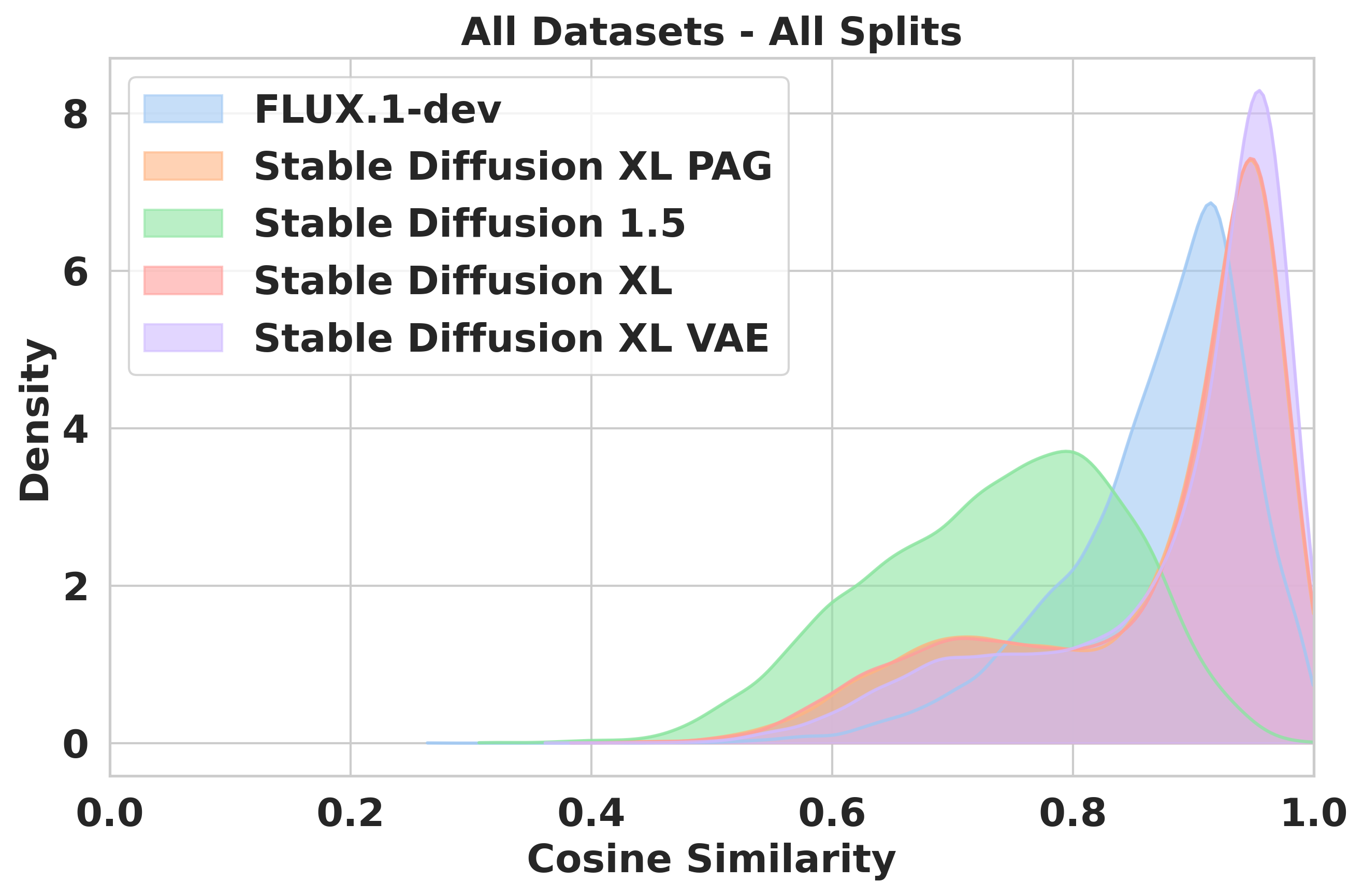}
  \caption{KDE distribution of image similarity between real and generated images across five image models.}
  \label{fig:image-density}
\end{figure}

For image models, we calculated both the Fréchet Inception Distance (FID)~\cite{Seitzer2020FID} and the average cosine image similarity between AI-generated images and collected images. FID is a common metric used to evaluate the quality of images created from generative models by comparing the distribution of generated images with that of real images. A lower FID score indicates greater similarity between the generated and real images. The average cosine similarity is computed between the vector representations of AI and real images, extracted using ResNet~\cite{resnet}. The average FID and average cosine image similarities are reported in right side of Table~\ref{tab:side-by-side}, we also plot KDE of the cosine image similarity in Figure~\ref{fig:image-density}. We observe that Stable Diffusion 1.5 exhibits the greatest image variance compared to other models, while Stable Diffusion XL and its VAE and PAG variations generate the most similar images. These trends hold for both the average and distribution analyses of the image models. Unlike the text models, no generated images exhibit low similarity when compared to real images.

\section{Experiments}
Existing multimodal vision LLMs can provide insightful descriptions for input text and image data, offering the most accessible way to interact compared to other deep learning models that require extensive training. We utilise both proprietary and open-source models as baselines to detect data reliability and originality using our MM-Health dataset and describe the benchmarking of these models.

\begin{table*}[]
\centering
\resizebox{\textwidth}{!}{%
\begin{tabular}{ccccccccccccccccccccccc}
\midrule
                                                                         &           & \multicolumn{10}{c}{\textbf{Reliable}}                                                                                                                                                            & \textbf{} & \multicolumn{10}{c}{\textbf{Unreliable}}                                                                                                                                                          \\ \cmidrule{3-12} \cmidrule{14-23} 
                                                                         &           & \textbf{4o} & \textbf{4o-mini} & \multicolumn{2}{c}{\textbf{Llama3.2-V}} & \multicolumn{2}{c}{\textbf{LLaVA-1.6}} & \multicolumn{2}{c}{\textbf{Qwen2-VL}} & \multicolumn{2}{c}{\textbf{MedGemma}} & \textbf{} & \textbf{4o} & \textbf{4o-mini} & \multicolumn{2}{c}{\textbf{Llama3.2-V}} & \multicolumn{2}{c}{\textbf{LLaVA-1.5}} & \multicolumn{2}{c}{\textbf{Qwen2-VL}} & \multicolumn{2}{c}{\textbf{MedGemma}} \\ \cmidrule{3-23} 
                                                                         &           & ZS          & ZS               & ZS                 & FS                 & ZS                 & FS                & ZS                & FS                & ZS                & FS                &           & ZS          & ZS               & ZS                 & FS                 & ZS                 & FS                & ZS                & FS                & ZS                & FS                \\ \midrule
\multirow{2}{*}{Text}                                                    & Human     & 0.499       & 0.499            & 0.500              & 0.499              & 0.500              & 0.494             & 0.499             & 0.499             & 0.499             & 0.485             &           & 0.334       & 0.328            & 0.246              & 0.376              & 0.077              & 0.348             & 0.162             & 0.348             & 0.295             & 0.409             \\
                                                                         & AI        & 0.499       & 0.500            & 1.000              & 0.436              & 1.000              & 0.498             & 0.500             & 0.499             & 1.000             & 0.499             &           & 0.183       & 0.105            & 0.020              & 0.238              & 0.010              & 0.101             & 0.019             & 0.113             & 0.081             & 0.208             \\ \midrule
\multirow{8}{*}{\begin{tabular}[c]{@{}c@{}}Text \&\\ Image\end{tabular}} & Human     & 0.499       & 0.499            & 0.499              & 0.499              & 0.500              & 0.495             & 0.499             & 0.498             & 0.527             & 0.495             &           & 0.371       & 0.353            & 0.312              & 0.373              & 0.085              & 0.351             & 0.140             & 0.340             & 0.304             & 0.393             \\
                                                                         & AI        & 0.498       & 0.499            & 0.498              & 0.454              & 1.000              & 0.498             & 0.499             & 0.500             & 0.532             & 0.499             &           & 0.314       & 0.261            & 0.250              & 0.331              & 0.030              & 0.207             & 0.084             & 0.247             & 0.229             & 0.327             \\ \cmidrule{2-23} 
                                                                         & AI-T 25\% & 0.499       & 0.499            & 0.499              & 0.468              & 1.000              & 0.497             & 0.498             & 0.500             & 0.499             & 0.499             &           & 0.347       & 0.304            & 0.287              & 0.349              & 0.065              & 0.267             & 0.124             & 0.302             & 0.274             & 0.381             \\
                                                                         & AI-T 50\% & 0.500       & 1.000            & 0.499              & 0.477              & 1.000              & 0.497             & 0.499             & 0.500             & 0.499             & 0.499             &           & 0.347       & 0.304            & 0.286              & 0.347              & 0.065              & 0.267             & 0.124             & 0.302             & 0.274             & 0.385             \\
                                                                         & AI-T 75\% & 1.000       & 1.000            & 1.000              & 0.490              & 1.000              & 0.497             & 0.499             & 0.500             & 0.499             & 0.499             &           & 0.395       & 0.372            & 0.330              & 0.381              & 0.086              & 0.339             & 0.207             & 0.392             & 0.363             & 0.437             \\ \cmidrule{2-23} 
                                                                         & AI-I 25\% & 0.498       & 0.499            & 0.498              & 0.450              & 1.000              & 0.499             & 0.500             & 0.500             & 0.499             & 0.499             &           & 0.313       & 0.262            & 0.245              & 0.328              & 0.040              & 0.221             & 0.089             & 0.244             & 0.231             & 0.333             \\
                                                                         & AI-I 50\% & 0.499       & 1.000            & 0.499              & 0.450              & 1.000              & 0.499             & 0.500             & 0.499             & 0.500             & 0.500             &           & 0.310       & 0.264            & 0.253              & 0.333              & 0.043              & 0.219             & 0.075             & 0.231             & 0.261             & 0.341             \\
                                                                         & AI-I 75\% & 1.000       & 1.000            & 0.499              & 0.449              & 1.000              & 0.497             & 1.000             & 0.499             & 1.000             & 1.000             &           & 0.310       & 0.265            & 0.252              & 0.333              & 0.037              & 0.212             & 0.077             & 0.244             & 0.253             & 0.336             \\ \midrule
\end{tabular}
}
\caption{Marco F1 score of \textbf{Task 1 information reliability check}. "ZS" is zero-shot and "FS" is five-shot.}
\label{tab:task1}
\end{table*}

\subsection{Task Definition}
We designed three benchmark tasks to evaluate baseline models' performance against both the reliability and originality of the health information collected in MM-Health. Those tasks are described as follows:

\noindent \textbf{Task 1: Information Reliability Check.} This task evaluates the VLLMs' capability to distinguish whether the provided information is reliable or unreliable. We provide the models with three different data settings: text-only data with human and AI content separated, text-image data with human and AI content separated, and text-image data with human and AI content mixed.

\noindent \textbf{Task 2: Information Originality Check.} This task evaluates the VLLMs' capability to distinguish whether the provided information is AI-generated, human-generated, or mixed. Following a similar approach to Task 1, we provide the models with content that is either separated into human and AI categories or mixed.

\noindent \textbf{Task 3: Fine-Grained AI Detection Analysis.} This task provides deeper insights into the information originality check from Task 2. Since we applied five different text and image generation models for our MM-Health dataset, we mixed the AI-generated data to create twenty-five different variations. This allows us to evaluate which combinations are the easiest or hardest for baseline models to detect as AI-generated.

\subsection{Baseline Models}
We established various multimodal baselines, including both proprietary and open-source models for a comprehensive comparison. \textbf{Proprietary models:} We employed GPT-4o and GPT-4o Mini, which have demonstrated excellent performance on open-source benchmarks such as MMLU~\cite{mmmlu}, HumanEval~\cite{chen2021codex}, and MMMU~\cite{MMMU}, and are widely adopted in vision and language research. These models serve as a benchmark standard, representing the top zero-shot performance that current state-of-the-art VLLMs can achieve. \textbf{Open-sourced models:} We include three general open-source VLLMs for a comprehensive comparison. These models are: Llama-3.2-Vision, built on the Llama-3.1-8B text model with a 2-billion-parameter vision encoder; LLaVA-1.6, which utilises the 7-billion-parameter Qwen v1.5 model with CLIP-ViT-L-336 attached via an MLP projection layer; and Qwen2-VL, which employs Multimodal Rotary Position Embedding (M-RoPE) with a 675M vision encoder and Qwen2-7B, resulting in a total of 7.6B parameters. To further access the performance of domain specific model, we also include MedGemma, which is a 4B model of Gemma3 variants that are trained for performance on medical text and image comprehension. These models demonstrate competitive zero-shot performance on vision-and-language benchmarks. 

\subsection{Experiment Settings.} For the GPT-4o and GPT-4o mini models, we use the OpenAI API endpoints: \texttt{gpt-4o-2024-08-06} and \texttt{gpt-4o-mini-2024-07-18}, respectively, throughout our experiments. Open-source models are hosted on a server equipped with an Nvidia A6000 GPU with 48GB of VRAM. To replicate realistic scenarios in most cases, we retain the default parameter settings for all models. To fully utilise the potential of the models, we experiment with baseline models using both zero-shot and five-shot settings, employing consistent prompts designed for different tasks. Zero-shot is directly evaluated on the testing set, while five-shot uses five samples from the training set and evaluates the models on the testing set. We use standard macro F1 scores to evaluate the baseline models across all three tasks. See appendix for details.

\section{Result}
\begin{table*}[]
\centering
\resizebox{\textwidth}{!}{%
\begin{tabular}{ccccccccccccccccccccccc}
\midrule
\textbf{}                                                                         & \textbf{}          & \multicolumn{10}{c}{\textbf{Reliable}}                                                                                                                                                            &  & \multicolumn{10}{c}{\textbf{Unreliable}}                                                                                                                                                          \\ \cmidrule{3-12} \cmidrule{14-23} 
\textbf{}                                                                         & \textbf{}          & \textbf{4o} & \textbf{4o-mini} & \multicolumn{2}{c}{\textbf{Llama3.2-V}} & \multicolumn{2}{c}{\textbf{LLaVA-1.6}} & \multicolumn{2}{c}{\textbf{Qwen2-VL}} & \multicolumn{2}{c}{\textbf{MedGemma}} &  & \textbf{4o} & \textbf{4o-mini} & \multicolumn{2}{c}{\textbf{Llama3.2-V}} & \multicolumn{2}{c}{\textbf{LLaVA-1.6}} & \multicolumn{2}{c}{\textbf{Qwen2-VL}} & \multicolumn{2}{l}{\textbf{MedGemma}} \\ \cmidrule{3-23} 
\textbf{}                                                                         & \textbf{}          & \textbf{ZS} & \textbf{ZS}      & \textbf{ZS}        & \textbf{FS}        & \textbf{ZS}        & \textbf{FS}       & \textbf{ZS}       & \textbf{FS}       & \textbf{ZS}       & \textbf{FS}       &  & \textbf{ZS} & \textbf{ZS}      & \textbf{ZS}        & \textbf{FS}        & \textbf{ZS}        & \textbf{FS}       & \textbf{ZS}       & \textbf{FS}       & \textbf{ZS}       & \textbf{FS}       \\ \midrule
\multirow{2}{*}{\textbf{Text}}                                                    & \textbf{Human}     & 0.316       & 0.323            & 0.023              & 0.321              & 0.242              & 0.321             & 0.493             & 0.324             & 0.001             & 0.075             &  & 0.254       & 0.266            & 0.012              & 0.255              & 0.201              & 0.267             & 0.320             & 0.288             & 0.006             & 0.023             \\
                                                                                  & \textbf{AI}        & 0.093       & 0.074            & 0.327              & 0.057              & 0.211              & 0.041             & 0.003             & 0.046             & 0.000             & 0.166             &  & 0.113       & 0.122            & 0.495              & 0.083              & 0.220              & 0.062             & 0.003             & 0.043             & 0.028             & 0.203             \\ \midrule
\multirow{8}{*}{\textbf{\begin{tabular}[c]{@{}c@{}}Text \&\\ Image\end{tabular}}} & \textbf{Human}     & 0.234       & 0.304            & 0.140              & 0.312              & 0.137              & 0.319             & 0.321             & 0.317             & 0.008             & 0.012             &  & 0.263       & 0.239            & 0.069              & 0.237              & 0.138              & 0.279             & 0.301             & 0.298             & 0.009             & 0.014             \\
                                                                                  & \textbf{AI}        & 0.155       & 0.119            & 0.188              & 0.122              & 0.271              & 0.070             & 0.093             & 0.071             & 0.156             & 0.174             &  & 0.215       & 0.233            & 0.231              & 0.164              & 0.282              & 0.063             & 0.091             & 0.099             & 0.211             & 0.178             \\ \cmidrule{2-23} 
                                                                                  & \textbf{AI-T 25\%} & 0.234       & 0.172            & 0.261              & 0.144              & 0.261              & 0.091             & 0.122             & 0.094             & 0.199             & 0.215             &  & 0.285       & 0.335            & 0.259              & 0.145              & 0.272              & 0.079             & 0.122             & 0.089             & 0.279             & 0.241             \\
                                                                                  & \textbf{AI-T 50\%} & 0.268       & 0.188            & 0.316              & 0.173              & 0.239              & 0.118             & 0.143             & 0.125             & 0.253             & 0.283             &  & 0.306       & 0.339            & 0.284              & 0.159              & 0.270              & 0.091             & 0.111             & 0.097             & 0.308             & 0.267             \\
                                                                                  & \textbf{AI-T 75\%} & 0.307       & 0.205            & 0.338              & 0.171              & 0.203              & 0.127             & 0.162             & 0.138             & 0.280             & 0.309             &  & 0.389       & 0.400            & 0.305              & 0.221              & 0.173              & 0.096             & 0.134             & 0.116             & 0.272             & 0.262             \\ \cmidrule{2-23} 
                                                                                  & \textbf{AI-I 25\%} & 0.236       & 0.143            & 0.256              & 0.173              & 0.262              & 0.100             & 0.114             & 0.096             & 0.207             & 0.218             &  & 0.261       & 0.264            & 0.279              & 0.176              & 0.295              & 0.067             & 0.093             & 0.118             & 0.262             & 0.235             \\
                                                                                  & \textbf{AI-I 50\%} & 0.270       & 0.180            & 0.284              & 0.215              & 0.235              & 0.109             & 0.143             & 0.108             & 0.246             & 0.256             &  & 0.369       & 0.282            & 0.319              & 0.253              & 0.193              & 0.082             & 0.117             & 0.188             & 0.259             & 0.276             \\
                                                                                  & \textbf{AI-I 75\%} & 0.322       & 0.218            & 0.333              & 0.246              & 0.185              & 0.134             & 0.156             & 0.130             & 0.291             & 0.283             &  & 0.358       & 0.275            & 0.316              & 0.203              & 0.154              & 0.093             & 0.118             & 0.194             & 0.266             & 0.294             \\ \midrule
\end{tabular}
}
\caption{Marco F1 score of \textbf{Task 2 information originality check}. "ZS" is zero-shot and "FS" is five-shot.}
\label{tab:task2}
\end{table*}

\begin{figure*}[h!]
  \centering
  \includegraphics[width=0.85\textwidth]{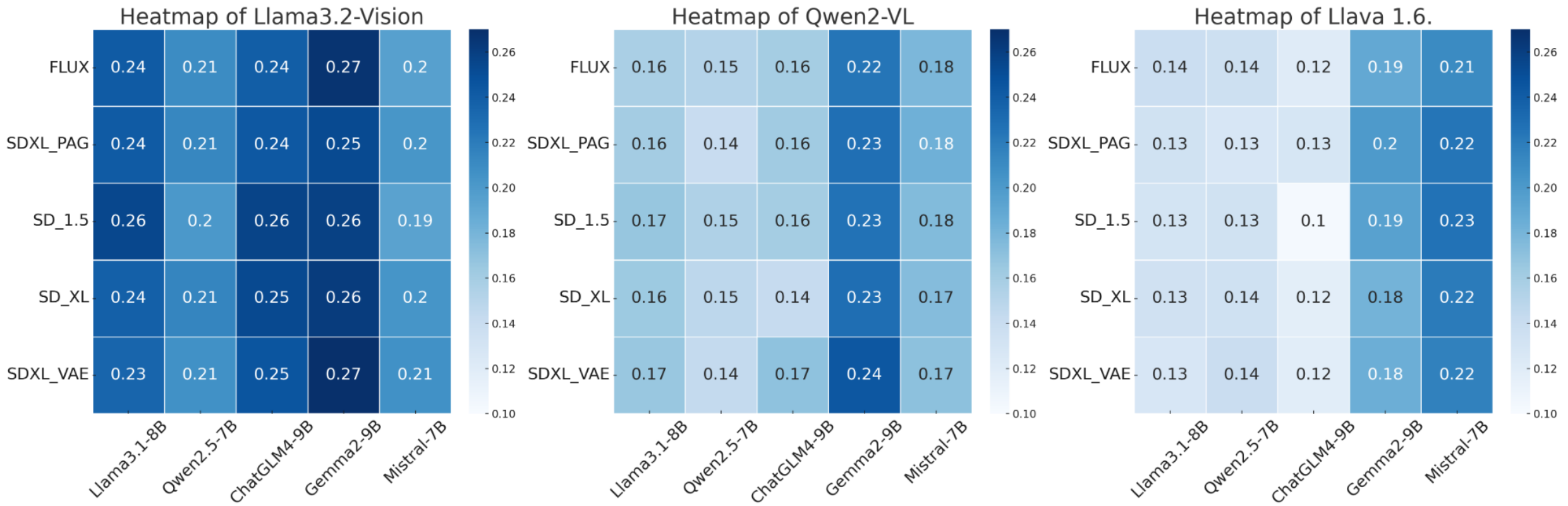}
  \caption{Heatmap representation of the \textbf{ Task 3 fine-grained AI detection analysis}. Each heatmap illustrates F1 scores from various VLLMs across twenty-five different combinations of AI-generated content. Darker colours represent higher F1 scores.}
  \label{fig:heat-map}
\end{figure*}

\noindent \textbf{Task 1: Information Reliability Check.} The results of the information reliability check are presented in Table~\ref{tab:task1}. We divide the dataset into reliable and unreliable articles and apply VLLMs for binary classification across various data settings, including text-only, text-image, and mixed text-image content incorporating both human and AI-generated inputs. To assess the impact of AI-generated content, we introduce varying proportions of AI-generated context at 25\%, 50\%, and 75\%, in both text and image modalities. Baseline models perform better at classifying reliable articles than unreliable ones, with a significant disparity in F1 scores observed in both zero-shot and five-shot settings. In zero-shot setting, some models, such as Llava-1.6, MedGemma, GPT-4o, and GPT-4o-min, achieve perfect F1 scores for reliable articles but struggle to classify unreliable articles. This may be caused by that the models tend to classify input information as reliable without proper verification. In the five-shot setting, models' performance improve on unreliable articles but worsen on reliable ones. This suggests that example-based learning helps the models better understand content for verification. MedGemma's noticeably higher zero-shot and five-shot performance than other open-source models suggests that a model with health-domain-specific knowledge is better equipped to identify the reliability of health-related information.

\noindent \textbf{Task 2: Information Originality Check.} The results of the originality check are presented in Table~\ref{tab:task2}. We classify the dataset into three categories: human-generated, AI-generated, and a mix of text-image content combining human and AI inputs, for multi-label classification. VLLMs are tasked with distinguishing whether the input context is human, AI, or human-AI mixed. Similar to Task 1, we introduce AI-generated content in both text and images at proportions of 25\%, 50\%, and 75\% to further evaluate the robustness of VLLMs in distinguishing mixed content. Baseline models struggle to accurately classify the originality of articles in both zero-shot and five-shot settings. We observe that baseline models may perform worse in the five-shot setting, suggesting that current VLLMs may lack the ability to learn comprehensive representations for assessing information originality using examples. GPT-4o performs the best among all the models, with MedGemma ranking as the second-best model, slightly trailing GPT-4o in performance. This performance difference is likely attributed to the larger dataset and model size used in GPT-4o’s training as well as health specific fine-tuning in MedGemma training.

\noindent \textbf{Task 3: Fine-grained AI Detection Analysis.} We visualise the performance of VLLMs using heatmaps that display the combinations of five text and five image models, resulting in a total of twenty-five results per heatmap, as shown in Figure~\ref{fig:heat-map}. Each heatmap displays the F1 scores of different open-source VLLMs in the five-shot setting, with darker colours indicating higher F1 scores in the binary classification of AI-generated content. A comparison between the GPT-4o and GPT-4o-mini models in the zero-shot setting is provided in the appendix Figure~\ref{fig:task3-heatmap-gpt} and Figure~\ref{fig:task3-heatmap-open}. Among the open-source models, Llama-3.2-Vision performs the best, followed by Qwen2-VL and Llava-1.6. We observe that text generated by Gemma-2-9B is more likely to be detected, while text generated by Qwen-2.5-7B is less likely to be detected. However, no consistent trend is observed for image models. The average F1 scores across all baseline models are approximately 0.2, indicating that none of the AI models cannot be reliably detected.

The results across all three tasks highlight the challenges faced by existing SOTA VLLMs. For information reliability check, model exhibit a strong bias towards reliable articles but struggle with unreliable content. In the information originality check, models fail to effectively differentiate between human, AI and mixed content, with five-shot setting showing no significant improvement. The fine-grained AI detection analysis decompose the model performance towards different generative models, with baseline models achieving low average F1 scores, indicating the need for more advanced detection methods. These findings demonstrate that MM-Health presents a challenging benchmark for the community. By releasing the MM-Health dataset, we aim to encourage the development of more robust and effective models to accurately classifying multimodal health misinformation.

\section{Case Study}

\subsection{Task 1: Information reliability check} 
Models demonstrated a tendency to rely on superficial stylistic cues rather than semantic content, leading to predictable errors in both directions. 

\noindent \textbf{Unreliable Content Misclassified as Reliable.} Models were consistently deceived by unreliable articles that mimicked credible journalism. These articles successfully evaded detection by anchoring misleading narratives to a verifiable fact, adopting a professional tone, and avoiding obvious red flags. Instead of containing clear falsehoods, they relied on biased framing and strategic omissions, which the models failed to identify.

\noindent \textbf{Reliable Content Misclassified as Unreliable.} Conversely, factually accurate content was often misclassified as unreliable, primarily due to a lack of explicit citations. Models also applied overly cautious heuristics to sensitive domains like medicine and politics, incorrectly penalizing content for using generalized language or offering reasonable predictive analysis.

\subsection{Task 2: Information originality check}
Similar patterns were observed in detecting text provenance, where models struggled with content that either perfectly emulated a known format or deviated from a perceived natural baseline.

\noindent \textbf{AI-Generated Content Misclassified as Human.} AI-generated text evaded detection by adhering closely to formal structures, such as news reports or scientific summaries. Its plausibility was enhanced by incorporating real-world details, like quoting public figures or referencing known events, causing its impersonal and structured format to be mistaken for authentic human writing.

\noindent \textbf{Human-Written Content Misclassified as AI.} Simultaneously, human-authored content was frequently misclassified as AI-generated. This was particularly common with text that was concise, logically structured, or satirical. The models appeared to equate clarity and logical structure with machine-generated text, and they failed to grasp rhetorical nuances like irony, often processing satirical arguments literally.

\section{Conclusion}
We introduce MM-Health, a multimodal dataset designed for detecting human and AI generated health misinformation. Unlike datasets limited to human generated information or augmented with LLM generated text, our approach combines text and image generative models to create multimodal counterparts of misinformation. To ensure diversity in generative model outputs, we incorporate five different open-source models for each modality. Our experiments involve six different VLLMs, including proprietary GPT-4o and GPT-4omini models, as well as SOTA open-source models, to evaluate data reliability and originality classification. Experimental results reveal that current VLLMs cannot accurately detect misinformation and AI-generated content, highlighting challenges in the era of generative AI. For future work, we plan to explore more robust detection techniques and expand the dataset to include other modalities such as video and audio. We hope publicly releasing the MM-Health dataset guides future research.

\section*{Acknowledgements}
This research was supported by the Macquarie University Research Acceleration Scheme (MQRAS) and Data Horizon funding.

\section*{Limitations}
In this work, we limited data collection to two modalities due to their prevalence on mainstream social platforms. This decision restricts the range of misinformation addressed by MM-Health, particularly from short-video platforms like TikTok, thereby limiting the scope of its potential applications. We encourage future studies to incorporate a broader variety of modalities to further enhance the diversity of machine-generated health information. Additionally, we exclusively employed out-of-the-box multimodal language models for misinformation detection, aiming to emulate common practices among the general public as our baseline. Given the existence of several specialised models designed for misinformation detection, we suggest further research to facilitate a more comprehensive comparison. Finally, there is a potential risk that the synthetic health misinformation could be misused by malicious actors. While we acknowledge this risk, the relatively small scale of the data generated in this study is unlikely to significantly exacerbate the existing challenge posed by the vast amount of health misinformation already circulating on the internet.

\section*{Ethical Considerations}
All experiments strictly adhere to the \href{https://www.aclweb.org/portal/content/acl-code-ethics%7D%7BACL}{Code of Ethics}. In Section \ref{sec:human_eval}, which details our human evaluation in data collection, we clearly informed the human evaluators of the task and that their responses would be utilized to assess the capabilities of large generative models. To ensure the anonymity and privacy of individuals involved in the data collection, we implemented a de-identification protocol. We directly remove all human evaluators' names associated with the generated data, all de-identified articles are stored in plain text format, without any identifying information. The original raw data are permanently deleted after the de-identification process. By taking these steps, we ensure that our data collection and analysis processes align with ethical guidelines and data protection regulations.

\bibliography{custom}

\clearpage
\appendix

\section{Remove Images}
Examples of removed images during health misinformation collection, those images include blurry, logo-based, fuzzy, or meaningless images, which are often downloaded during web scraping.

\begin{figure}[!h]
  \centering
  \includegraphics[width=0.4\textwidth]{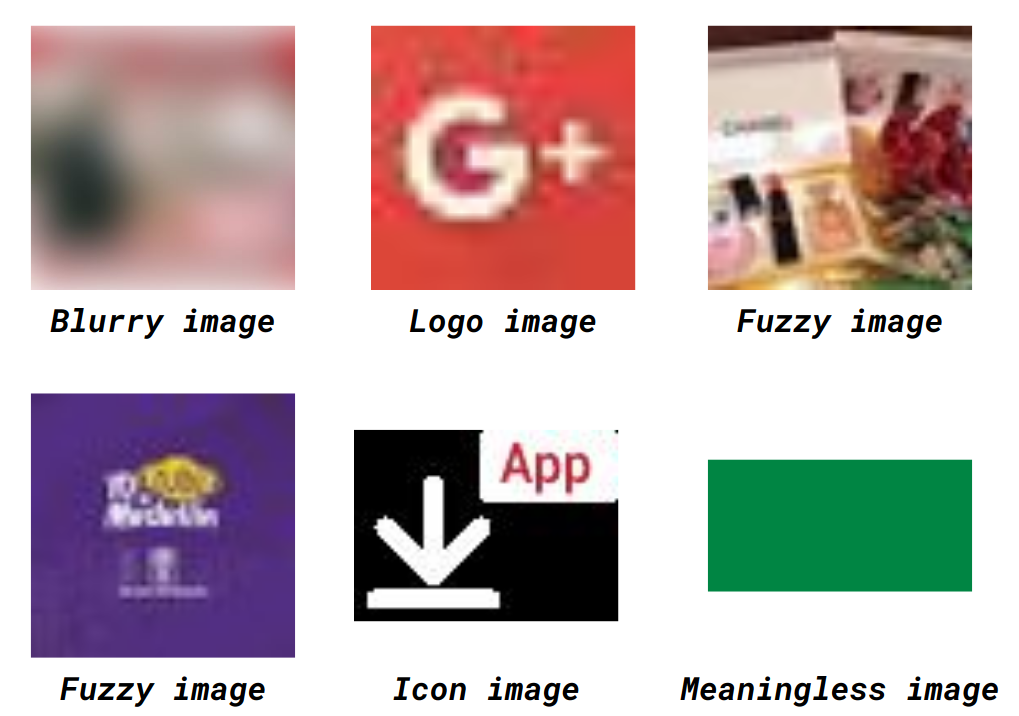}
  \caption{Sample of removed images after web scrapping. These images are irrelevant to the health topic, including blurry, logo-based, fuzzy, or meaningless images.}
  \label{fig:remove-images}
\end{figure}

\section{Overall Statistics}
The overall statics of MM-Health is presented in Table~\ref{tab:overall-statistic}, we utilised five different generative models for both text and image modalities to form the AI-generated portion of the dataset. The average text length and number of images are consistent across different sets.

\begin{table*}[]
\centering
\resizebox{0.8\textwidth}{!}{%
\begin{tabular}{@{}cccccclcccc@{}}
\toprule
\multirow{2}{*}{\textbf{Split}}   & \multirow{2}{*}{\textbf{Source}} & \multicolumn{4}{c}{\textbf{Reliable}}                                           &  & \multicolumn{4}{c}{\textbf{Unreliable}}                                         \\ \cmidrule(lr){3-6} \cmidrule(l){8-11} 
                                  &                                  & \textbf{Article \#} & \textbf{Avg Len.} & \textbf{Image \#} & \textbf{Avg IPA.} &  & \textbf{Article \#} & \textbf{Avg Len.} & \textbf{Image \#} & \textbf{Avg IPA.} \\ \midrule
\multirow{2}{*}{\textbf{Train}}   & \textbf{Human}                   & 3345                & 848.34            & 5175              & 1.55              &  & 809                 & 635.77            & 6917              & 8.55              \\
                                  & \textbf{AI}                      & 16725               & 463.84            & 25768             & 1.54              &  & 4045                & 450.56            & 34084             & 8.43              \\ \midrule
\multirow{2}{*}{\textbf{Val}}     & \textbf{Human}                   & 373                 & 856.01            & 586               & 1.57              &  & 90                  & 749.04            & 709               & 7.88              \\
                                  & \textbf{AI}                      & 1865                & 464.34            & 2917              & 1.56              &  & 450                 & 455.93            & 3415              & 7.59              \\ \midrule
\multirow{2}{*}{\textbf{Test}}    & \textbf{Human}                   & 932                 & 822.82            & 1435              & 1.54              &  & 227                 & 669.3             & 1572              & 6.93              \\
                                  & \textbf{AI}                      & 4660                & 460.21            & 7139              & 1.53              &  & 1135                & 450.28            & 7695              & 6.78              \\ \midrule
\multirow{2}{*}{\textbf{Overall}} & \textbf{Human}                   & 4650                & 843.84            & 7196              & 1.55              &  & 1126                & 651.58            & 9198              & 8.17              \\
                                  & \textbf{AI}                      & 23250               & 463.15            & 35824             & 1.54              &  & 5630                & 450.93            & 45194             & 8.03              \\ \bottomrule
\end{tabular}%
}
\caption{Overall statistics of the MM-Health dataset, where "Avg Len" represents the average word count of the articles, and "Avg IPA" indicates the average image count per article. The statistics are provided for both human and AI data across all data splits.}
\label{tab:overall-statistic}
\end{table*}

\section{Post-process Algorithm}
We developed Algorithm~\ref{alg:consensus_align} to post-process data alignment between the human-generated context and all five AI-generated contexts. $R_{text}$ and $R_{image}$ represent the human data collected from the open-source datasets, while $N_{text}^{(k)}$ and $N_{image}^{(k)}$ are inputs derived from AI-generated text and images, where $k$ ranges from one to five, representing the five different models. The outer loop iterates through each possible text-image pair across all models, while the inner loop checks the pair $N_{text_i}$ and $N_{image_j}$ against all five models. Only when text and image pairs from all five models exist will the matched data be added to $M_{final}$ to form the final dataset.

\begin{algorithm}[t]
\caption{Text and Image Alignment Across Models}\label{alg:consensus_align}
\begin{algorithmic}
\Ensure $R_{text}, R_{image}$ 
\Require $N_{text}^{(k)}, N_{image}^{(k)}, k=1,...,5$ 

\State $M_{final} \gets \emptyset$ 

\ForAll{$N_{text_i}, N_{image_j}$} 
    \State $match \gets \text{true}$ 
    
    \For{$k \gets 1$ to $5$} 
        \If{$N_{text_i}^{(k)} \notin R_{text} \lor N_{image_j}^{(k)} \notin R_{image}$}
            \State $match \gets \text{false}$
            \State \textbf{break} 
        \EndIf
    \EndFor
    
    \If{$match = \text{true}$} 
        \State $M_{final} \gets M_{final} \cup (R_{text_i}, R_{image_j}, N_{text_i}, N_{image_j})$
    \EndIf
\EndFor

\State \Return $M_{final}$ 
\end{algorithmic}
\end{algorithm}

\section{Generation Prompts}
\begin{figure*}[h]
  \centering
  \includegraphics[width=0.75\textwidth]{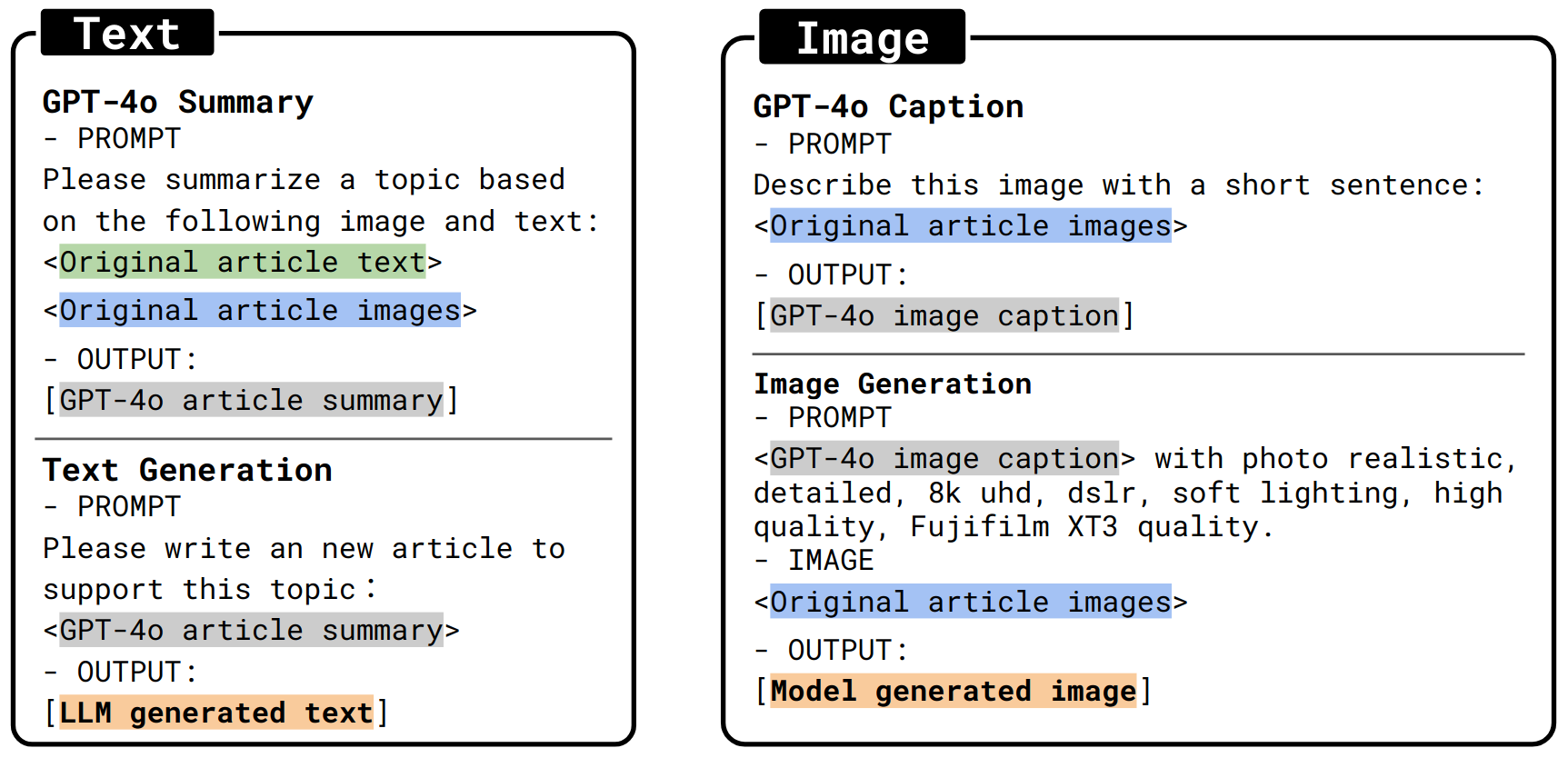}
  \caption{AI generation prompts. Text generation prompts are shown on the left side of the figure. Image generation prompts are shown on the right side of the figure.}
  \label{fig:prompts}
\end{figure*}

\begin{figure*}[!h]
    \centering
    \begin{subfigure}{\textwidth}
        \centering
        \includegraphics[width=0.9\textwidth]{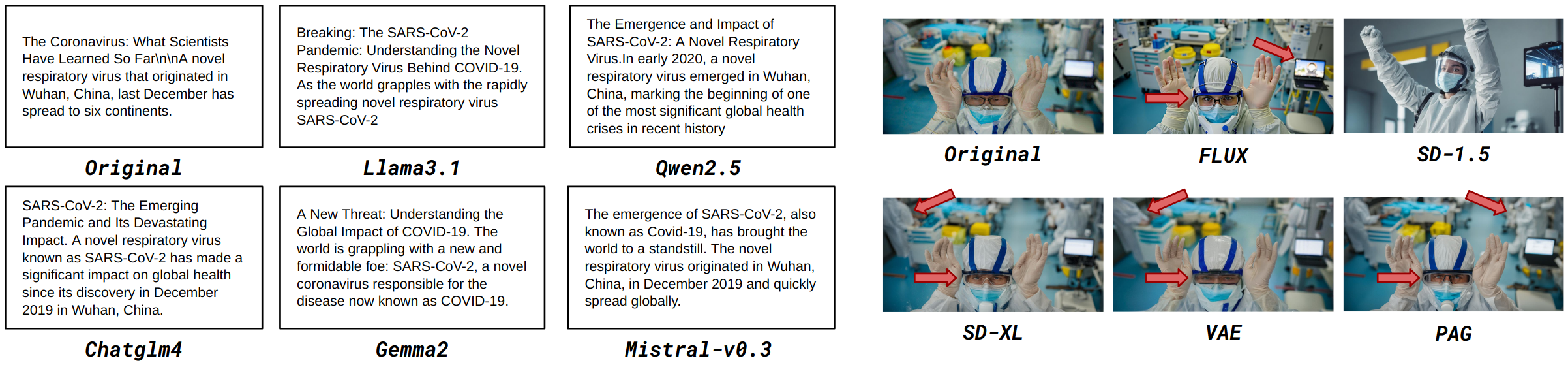}
        \caption{Reliable news generated from both human and AI.}
    \end{subfigure}
    \vspace{5pt} 
    \begin{subfigure}{\textwidth}
        \centering
        \includegraphics[width=0.9\textwidth]{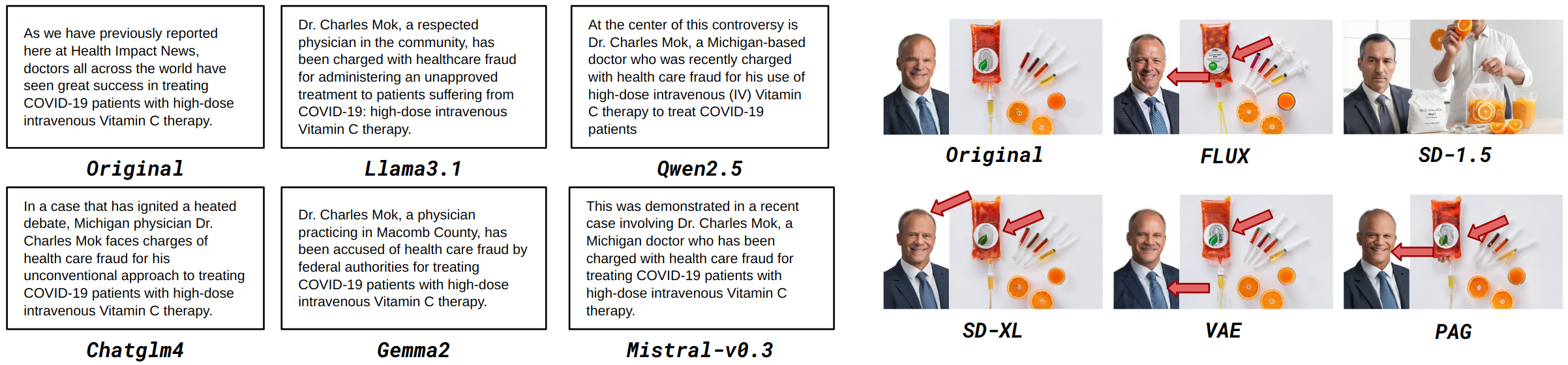}
        \caption{Unreliable news generated from both human and AI.}
    \end{subfigure}
    \caption{Examples of human and AI generated (a:) reliable and (b:) unreliable multimodal news. The examples at the top represent reliable news, while those at the bottom represent unreliable news. Note that the articles have been rewritten on the same topic and the image details have been manipulated, as pointed by the {\color{red} $\Rightarrow$}.}
    \label{fig:data-sample}
\end{figure*}

The generation prompts are presented in Figure~\ref{fig:prompts}. Within the prompts, text highlighted in blue represents the raw input images extracted from the news article. Green text highlights indicate the original news article content, while grey text shows the intermediate output from GPT-4o, and orange text denotes the final generated output. The generation examples are shown in Figure~\ref{fig:data-sample}, and we ensure the generations are similar to the original images but with modifications to mimic the real-world scenario where misinformation is only slightly different from the truth.

\begin{figure*}[!h]
  \centering
  \includegraphics[width=0.9\textwidth]{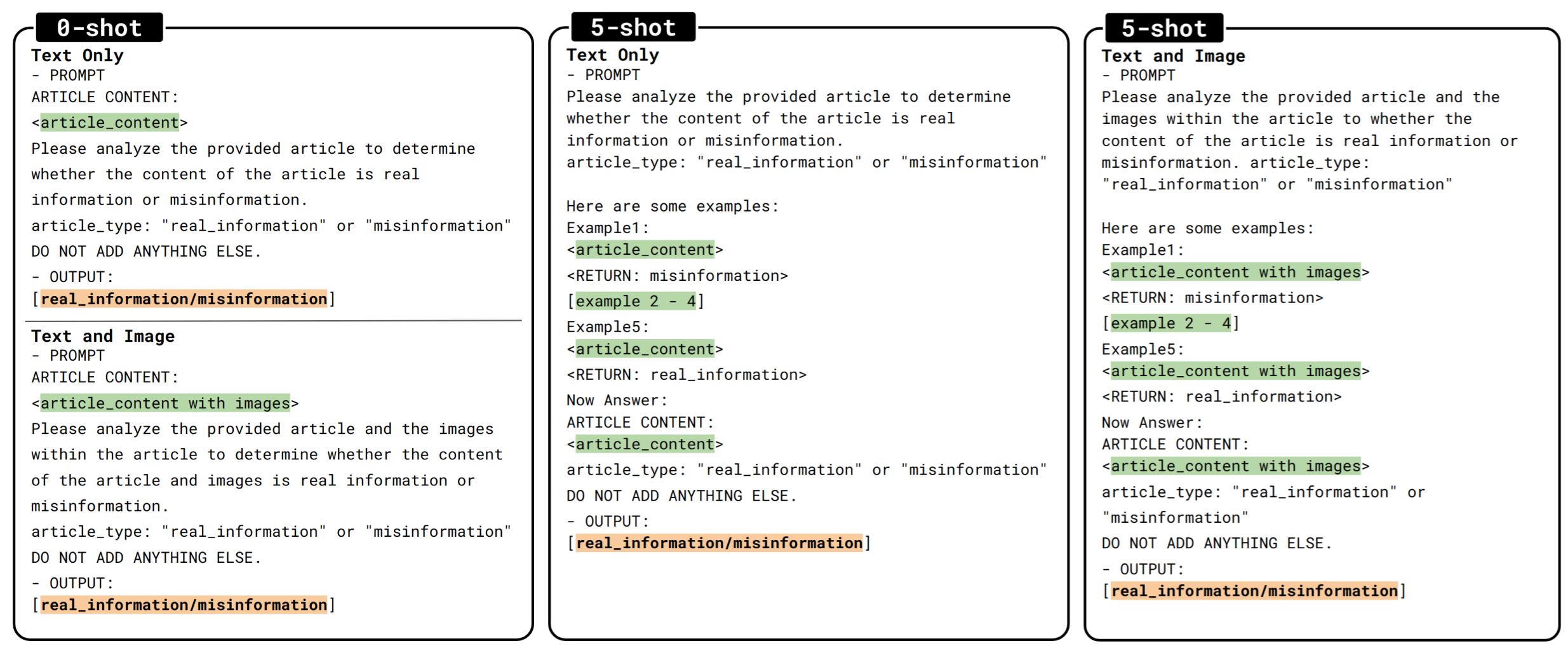}
  \caption{Zero-shot and five-shot prompts for task 1 - information reliability check.}
  \label{fig:task1-prompts}
\end{figure*}

\begin{figure*}[!h]
  \centering
  \includegraphics[width=0.9\textwidth]{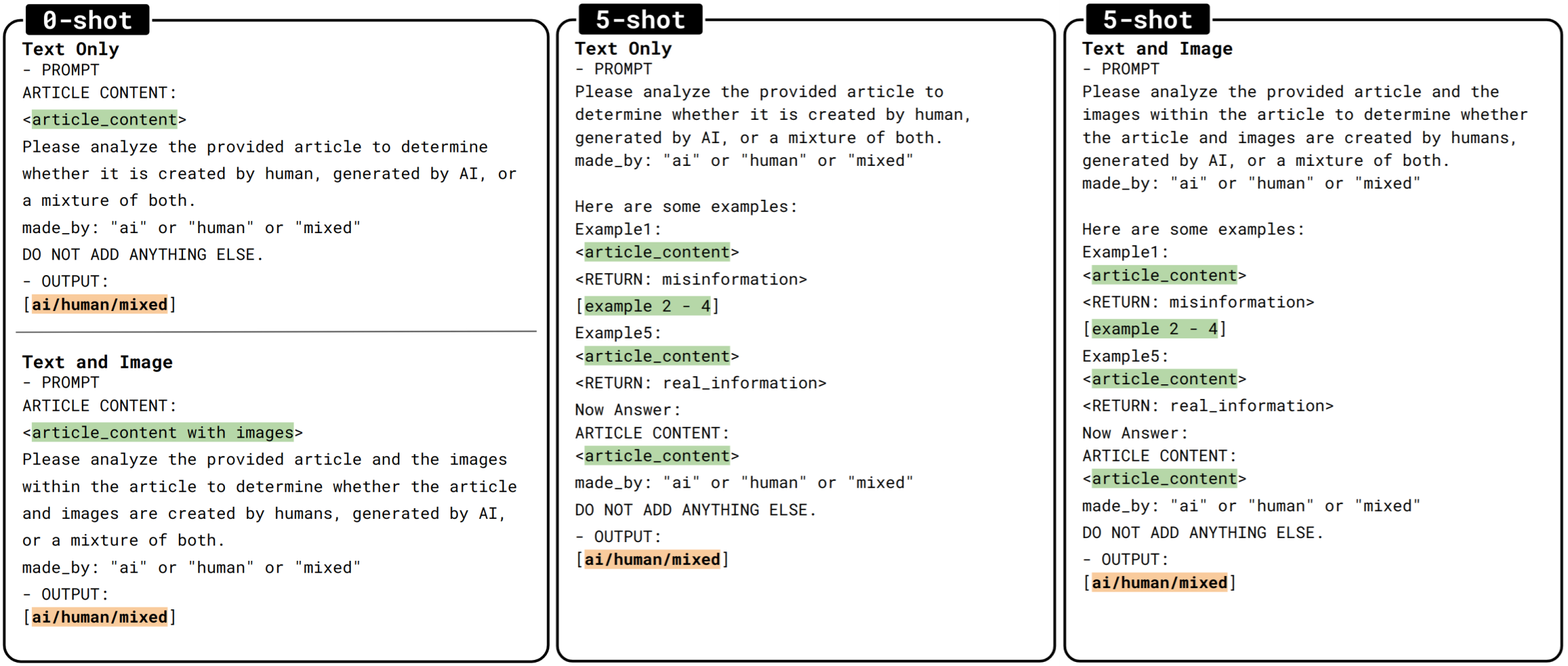}
  \caption{Zero-shot and five-shot prompts for task 2 - information originality check.}
  \label{fig:task2-prompts}
\end{figure*}

\section{Experiment Prompts}
\begin{figure*}[!h]
  \centering
  \includegraphics[width=0.73\textwidth]{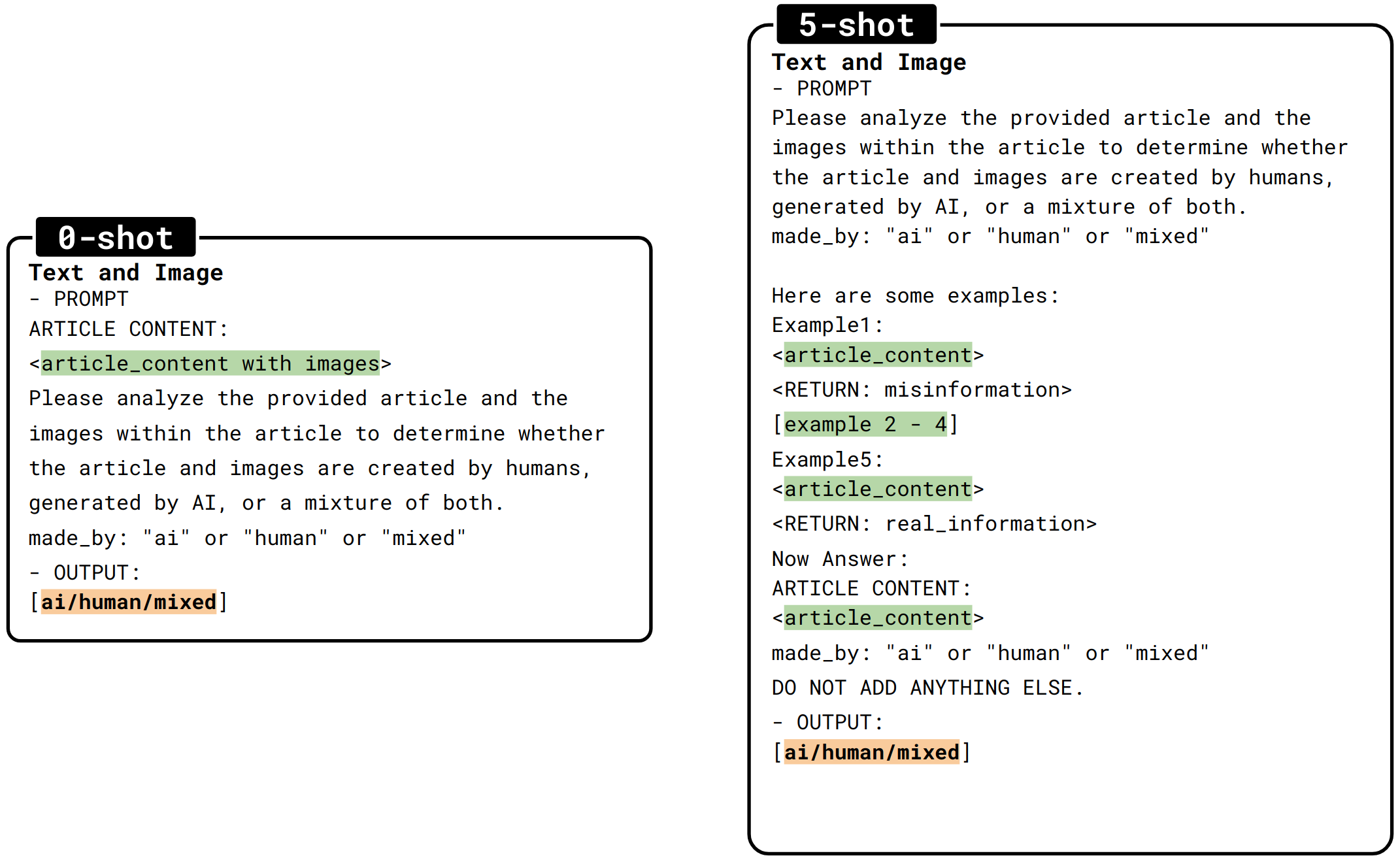}
  \caption{Zero-shot and five-shot prompts for task 3 - fine-grained AI detection analysis.}
  \label{fig:task3-prompts}
\end{figure*}

We present the prompts used for all three tasks in the experiment, covering both zero-shot and five-shot settings. Both proprietary and open-source models follow the same prompts. Figure~\ref{fig:task1-prompts} illustrates the zero-shot and five-shot prompts for the information reliability check, Figure~\ref{fig:task2-prompts} depicts the prompts for the information originality check, and Figure~\ref{fig:task3-prompts} presents the prompts for the fine-grained AI detection analysis. Article inputs are highlighted in green, while model outputs are highlighted in orange. Since models may not always produce the exact output specified in the prompts, we run each model several times until all responses are correctly formatted to meet the output requirements.

\section{Heatmaps}
\label{sec:heatmap}
\begin{figure*}[!h]
  \centering
  \includegraphics[width=0.73\textwidth]{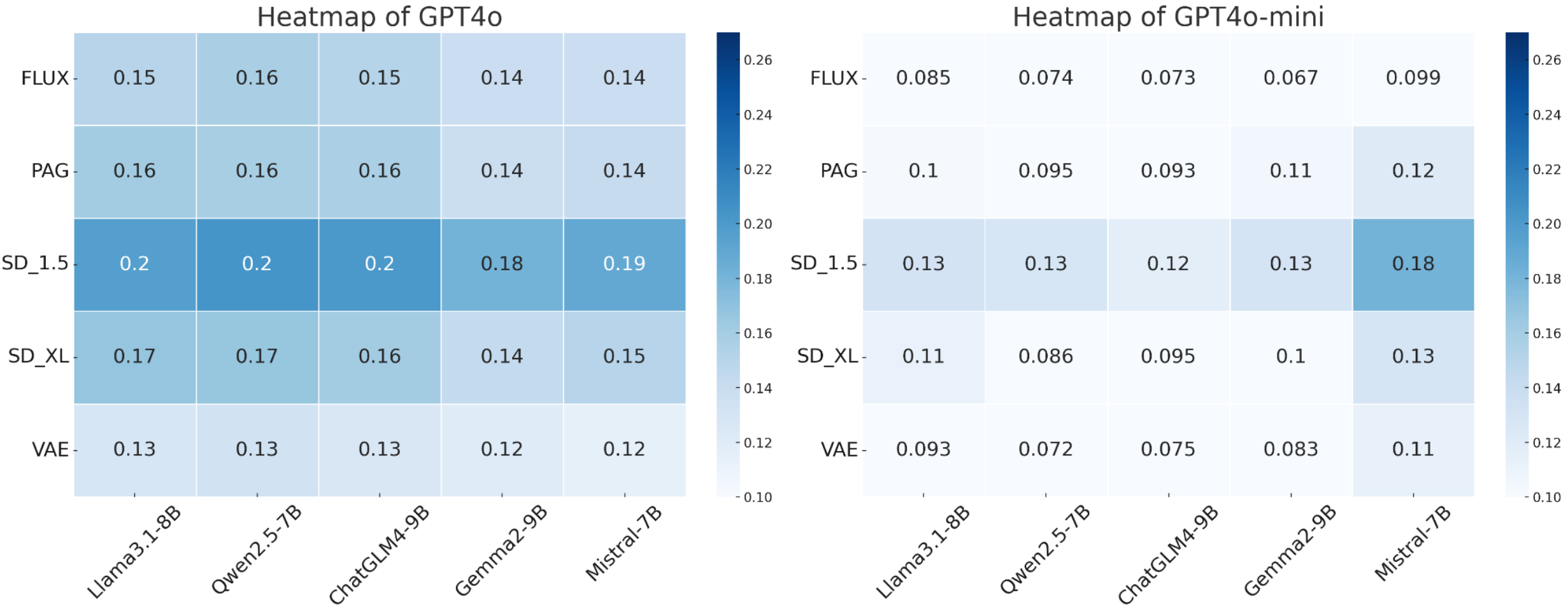}
  \caption{Zero-shot GPT4o and GPT4o-mini heatmaps for task 3 - fine-grained AI detection analysis.}
  \label{fig:task3-heatmap-gpt}
\end{figure*}

\begin{figure*}[]
  \centering
  \includegraphics[width=0.95\textwidth]{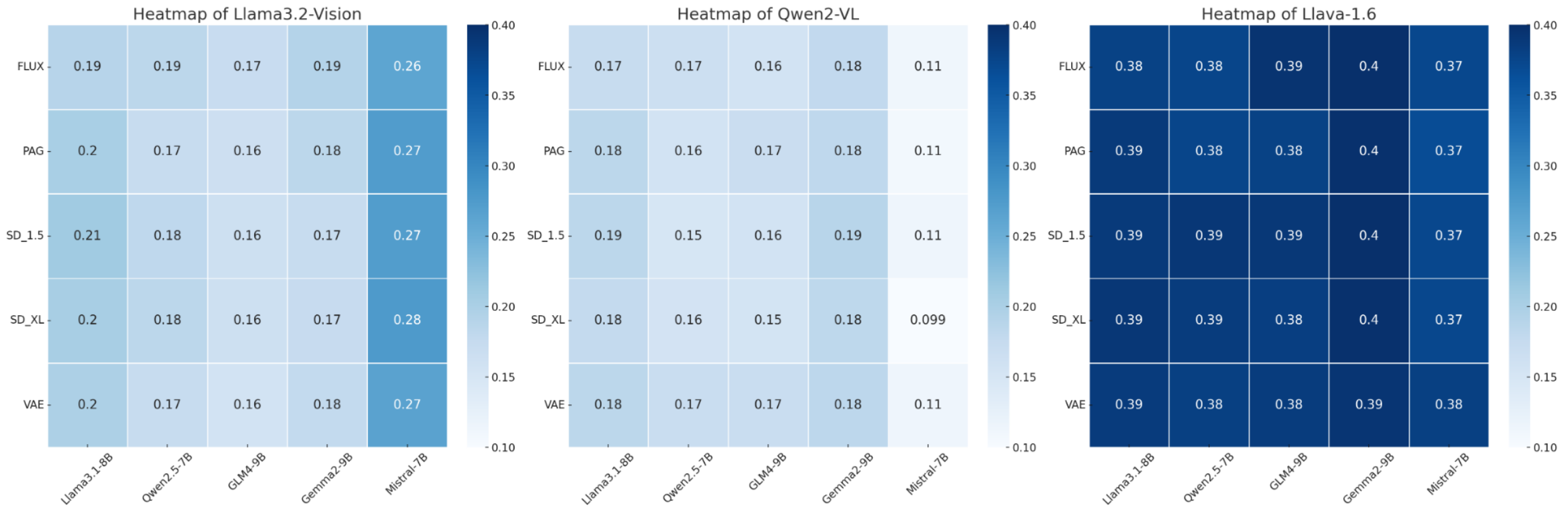}
  \caption{Zero-shot Llama-3.2-Vision, Qwen2-VL and Llava-1.6 heatmaps for task 3 - fine-grained AI detection analysis.}
  \label{fig:task3-heatmap-open}
\end{figure*}

We present the heatmaps in Figure~\ref{fig:task3-heatmap-gpt} and Figure~\ref{fig:task3-heatmap-open} for GPT4o, GPT4o-mini, Llama-3.2-Vision, Qwen2-VL and Llava-1.6 in zero-shot setting for task 3 - fine-grained AI detection analysis. All the models have been run through the exact zero-shot settings and prompts as demonstrated in Figure~\ref{fig:task3-prompts}.

\end{document}